\documentclass[10pt,twocolumn,letterpaper]{article}

\usepackage[pagenumbers]{cvpr} % To force page numbers, e.g. for an arXiv version

\usepackage[dvipsnames]{xcolor}

\definecolor{cvprblue}{rgb}{0.21,0.49,0.74}
\usepackage[pagebackref,breaklinks,colorlinks,citecolor=cvprblue]{hyperref}

\def\paperID{*****} % *** Enter the Paper ID here
\def\confName{CVPR}
\def\confYear{2024}

\title{Second Place Solution of WSDM2023 Toloka \\ Visual Question Answering Challenge}

\author{
Xiangyu Wu$^{1}$
\and
Zhouyang Chi$^{1}$
\and
Yang Yang$^{1}$
\and 
Jianfeng Lu$^{1}$
\and
$^1$ Nanjing University of Science and Technology \\
\{wxy\_yyjhl,1527612210,yyang,lujf\}@njust.edu.cn
}

\begin{document}
\maketitle
\begin{abstract}
In this paper, we present our solution for the WSDM2023 Toloka Visual Question Answering Challenge. Inspired by the application of multimodal pre-trained models to various downstream tasks(e.g., visual question answering, visual grounding, and cross-modal retrieval), we approached this competition as a visual grounding task, where the input is an image and a question, guiding the model to answer the question and display the answer as a bounding box on the image. We designed a three-stage solution for this task. Specifically, we used the visual-language pre-trained model OFA as the foundation. In the first stage, we constructed a large-scale synthetic dataset similar to the competition dataset and coarse-tuned the model to learn generalized semantic information. In the second stage, we treated the competition task as a visual grounding task, loaded the weights from the previous stage, and continued to fine-tune the model on the competition dataset, transferring the semantic information learned in the first stage to the competition task. Finally, we designed a bounding box matching and replacing post-processing strategy to correct the model's prediction results. Our team achieved a score of 76.342 on the final leaderboard, ranking second.
\end{abstract}
\section{Introduction}
\label{sec:intro}

Visual-language pretraining (VLP)~\cite{VLP-1,VLP-2,VLP-3,VLP-4,Paper-1,Paper-2} models have seen rapid development and significant advancements in recent years. These models aim to bridge the gap between visual and linguistic information, enabling machines to understand and generate contextually rich content combining images and text. Early work, such as OFA~\cite{ofa} and BLIP~\cite{BLIP}, laid the groundwork for integrating visual and textual information. OFA aimed to unify vision-language tasks using large-scale pretraining, while BLIP improved data efficiency and model performance through self-supervised learning. Recent advancements, including ChatGPT-4~\cite{GPT-4}, have led to more powerful models capable of tasks like image generation from text, multimodal conversation, and advanced visual reasoning.

One key application is Visual Question Answering (VQA)~\cite{VQA-1,VQA-2,VQA-3,VQA-4,Paper-3,Paper-4}, where the model answers questions based on an image. This task requires understanding both the image and the natural language question, combining image recognition, language processing, and reasoning. Visual grounding~\cite{VG-1,VG-2,VG-3,VG-4,VG-5,Paper-5} is a specific task within the broader domain of VQA. In VQA, the primary objective is to answer questions about an image using natural language understanding. Visual grounding within VQA focuses on precisely locating and identifying objects or regions in the image that correspond to specific elements mentioned in the textual query. This task requires the model to effectively link the textual description with relevant visual features in the image, enabling accurate and contextually relevant answers to questions posed about visual content. 

VLP models~\cite{VLP-1,VLP-2,Paper-6,Paper-7} achieve visual grounding tasks by first pretraining on large-scale datasets containing paired image and text data. These models extract feature embeddings from images using CNNs and process textual descriptions using transformer architectures like BERT. Through cross-modal attention mechanisms, they align visual and textual features, enabling precise linking of natural language queries to specific visual elements in images. Fine-tuning on task-specific datasets further refines their ability to understand and respond to queries that require nuanced visual comprehension within applications such as VQA and multimodal interaction systems.

With the rapid development of multimodal pre-trained models, they have shown strong generalization capabilities in various downstream tasks, such as visual grounding, visual question answering, cross-modal retrieval, image captioning, and more. We found that the visual grounding task refers to predicting the bounding box for a given text in an image, which inspired us to transform the competition task, namely visual question grounding, into a visual grounding task, effectively leveraging the power of multimodal pre-trained models. 

In the first stage, we collected and constructed a large amount of external data for coarse tuning of the model, with image distributions, question distributions, and label distributions similar to the competition data. Secondly, we used the OFA model as the foundation and coarse-tuned a model with strong generalization and rich semantics. In the second stage, we loaded the coarse-tuned weights from the previous stage, treated the competition task as a visual grounding task, constructed different question templates, and continued fine-tuning the OFA model on the competition dataset. Finally, we designed a bounding box matching and replacing the post-processing strategy. The bounding box position predicted by visual grounding is roughly correct, but the accuracy of coordinates is not higher than that of an object detector. Therefore, by calculating the IoU, we replaced the predicted bounding box with the object detector’s bounding box to correct the model's prediction results. Our team secured the runner-up position on the final leaderboard with a score of 76.342.

\section{Related Works}
\label{sec:formatting}

%-------------------------------------------------------------------------
\subsection{Visual-Language Pretraining (VLP)}

Recent developments in VLP~\cite{VLP-1,VLP-2,VLP-3,VLP-4,Paper-12,Paper-11} have been propelled by pioneering models such as OFA~\cite{ofa} and BLIP~\cite{BLIP}. These models have innovatively utilized large-scale datasets to pretrain transformer-based architectures on paired image-text data. OFA aims to unify various vision-language tasks by leveraging comprehensive pretraining, thereby enhancing performance in applications such as VQA and image-text retrieval. Meanwhile, BLIP focuses on optimizing data efficiency and model performance through advanced self-supervised learning techniques, contributing significantly to the robustness of multimodal understanding. In this evolving landscape, ChatGPT-4~\cite{GPT-4} has emerged as a notable advancement, showcasing remarkable capabilities in generating nuanced natural language responses and pushing the boundaries of language modeling tasks.

%-------------------------------------------------------------------------
\subsection{Visual Question Answering (VQA)}

VQA~\cite{VQA-1,VQA-2,VQA-3,VQA-4} involves answering questions about images using natural language understanding, which has been a pivotal area of research in multimodal AI. These tasks require models to comprehend both the visual content of images and the semantic context of textual queries. Early approaches focused on integrating vision and language through methods like joint embeddings and attention mechanisms. More recent advancements, leveraging models such as OFA~\cite{ofa} and BLIP~\cite{BLIP}, have significantly improved the accuracy and efficiency of VQA systems by employing large-scale pretraining on diverse datasets. These models aim to enhance the model's ability to reason about complex scenes and provide accurate responses based on multimodal inputs. VQA continues to evolve with the introduction of transformer-based architectures like ChatGPT-4, which further pushes the boundaries of multimodal understanding and response generation in AI systems.

%-------------------------------------------------------------------------
\subsection{Visual Grounding (VG)}
Visual grounding~\cite{VG-1,VG-2,VG-3,VG-4,VG-5,Paper-8} tasks involve the localization and identification of specific objects or regions within an image based on textual descriptions. This task is crucial in multimodal AI applications, where models need to link natural language queries to corresponding visual elements effectively. By leveraging deep learning techniques, such as convolutional neural networks (CNNs) for image feature extraction and transformer architectures for textual understanding, models can align visual and textual modalities. This alignment enables precise object localization, scene understanding, and context-based reasoning, enhancing applications like VQA~\cite{VQA-1,VQA-2}, image retrieval, and interactive systems that require accurate interpretation and response generation based on visual content.
\section{Methods}

The solution of this competition is based on the OFA~\cite{ofa} visual language pre-training model, so we introduce the OFA model first. OFA is a large multi-modal pre-training model that transforms all pre-training tasks into text generation tasks so that all pre-training tasks can be performed by one decoder module. For example, VG, visual grounding pre-training task construct the template which region does the text "Man in white shirt" describe? bounding box coordinates are generated by the decoder module; ITM, image text match pre-training task, construct the template, Does the image describe "Two boys playing frisbee on the grass"? yes or no is generated by the same decoder module. It can be seen, that the visual grounding task is very similar to the competition, which are both output bounding box coordinates. Therefore, we transformed the competition task into visual Grounding task and used the large pre-training model for fine-tuning.

Our methods mainly include three parts, respectively Coarse tuning, Fine tuning, and Postprocessing. In the first part, Coarse tuning aims to build a synthetic dataset similar to the competition dataset and train a weak semantics but strong generalization model on the synthetic dataset. The weak semantics is because the synthetic dataset contains dirty data and labels so the semantic learning is not accurate enough. Strong generalization is due to the large scale of the synthetic dataset, so the generalization ability is relatively strong. In the second part, the Fine-tuning stage loads the coarse-tuning part weight and continues to train the competition dataset. The third part, post-processing, through bounding box matching, replacing, and model ensemble, further improves the accuracy of model prediction.

\begin{figure}[!htbp]
	\centering
	\includegraphics[width=\linewidth]{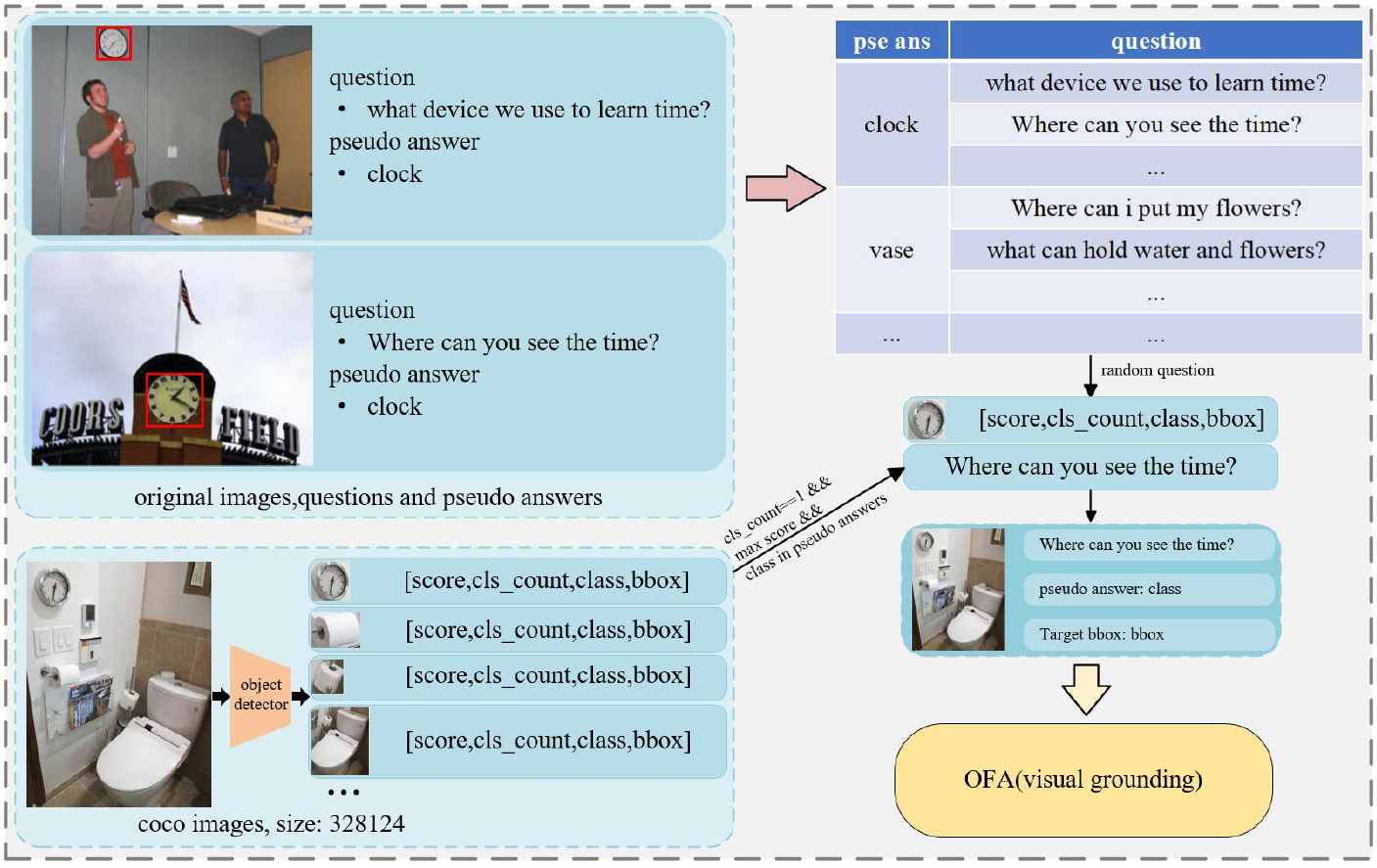}
	\caption{Coarse Tuning Stage.}
\label{fig: retrieval}
\end{figure}

\subsection{Coarse Tuning Stage}
The first part is the coarse tuning stage. We construct the coarse-tuning dataset through coco images. There are several features of constructing synthetic datasets. First, the image distribution, question distribution, and bounding box distribution of the coarse tuning dataset should be as close as possible to the competition dataset, so that the model can avoid learning the wrong data distribution; Secondly, any sampling operation must be Random Sampling, which can ensure that the model will not learn any data bias. Finally, do not sample any data in the test public set, which can avoid overfitting the model to the test public set. The coarse-tuning paradigm should be the same as the competition task, similar to the pre-training, save the coarse-tuned model weight for the next fine-tuning part. similar to the pre-training, the more coarse tuning epochs, the stronger generalization, and the better performance.

This figure illustrates our process of constructing the synthetic dataset. For each sample in the training set, we directly input images and questions into the multi-modal pre-training model to obtain textual answers, which we define as pseudo-answers. Now, each sample consists of an image, a question, and a pseudo-answer; therefore, we can build a mapping table of pseudo-answers and questions, where one pseudo-answer corresponds to multiple questions, for example, the pseudo-answers clock and vase. Then, randomly selected an image from the coco and detected several objects by the object detector. Each detected object includes confidence, class counts, class name, and bounding box, where class number is defined as the number of the object in the image, such as the number of clocks is one and the number of roll paper is two. Next, we sorted the detected objects according to confidence, selected an object class with the largest confidence, the class count is one, and belonging to the pseudo answer, then, randomly selected a question corresponding to the pseudo answer from the mapping table. Now, a new sample is obtained, the question of the sample is randomly selected question, the pseudo-answer is the selected object class, and the target bounding box is obtained by the object detector. After the synthetic dataset is constructed, a model with weak semantics but strong generalization can be coarse tuning in the form of visual grounding.

\begin{figure}[!htbp]
	\centering
	\includegraphics[width=\linewidth]{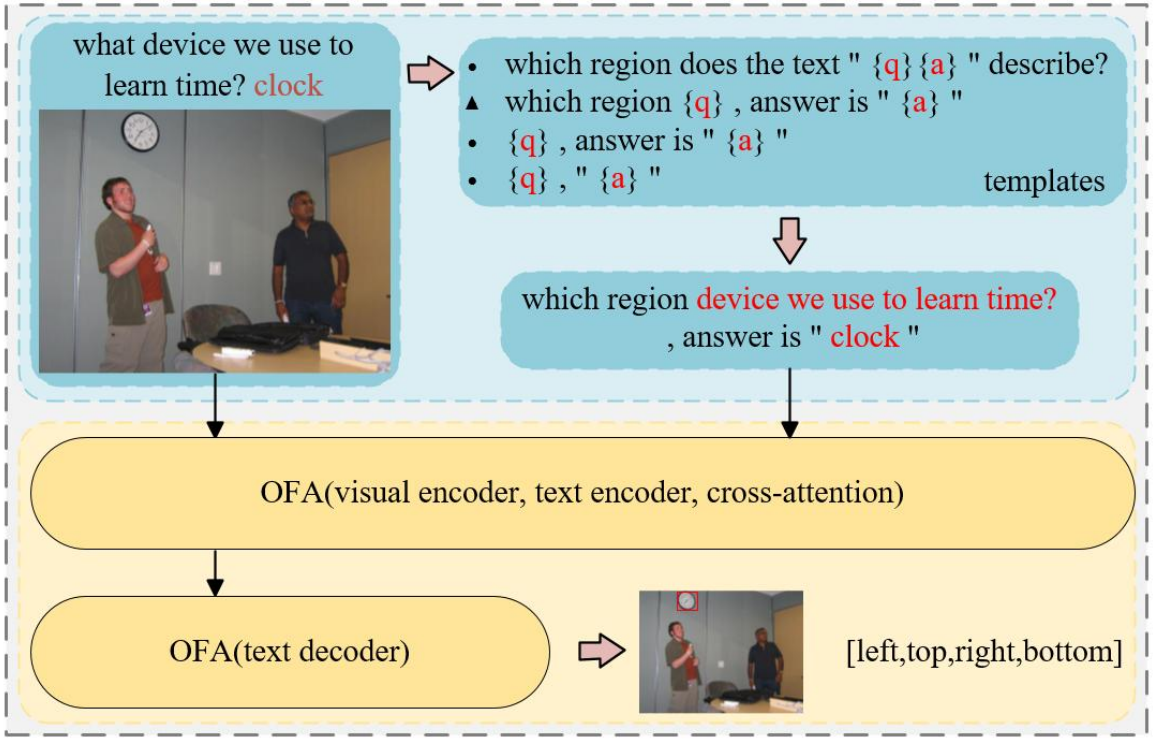}
	\caption{Fine Tuning Stage.}
\label{fig: similarity}
\end{figure}

\subsection{Fine-tuning Stage}
The second part is the Fine-tuning stage. In the form of a visual grounding task, load the trained weight obtained by the last Coarse tuning part and continue to train the model on the competition dataset. Each sample consists of an image, a question, and a pseudo-answer. Question and pseudo answer are combined into the template, which is used as text input of the model. For the four templates we constructed, we finally adopted the second template, and the first word of the question, such as what, where, and which, was replaced with which region. After the template construction, the template and image are input into the OFA model, through text encoder, image encoder, and cross-modal fusion, the coordinates of the bounding box are finally output by text decoder.

\begin{figure}[!htbp]
	\centering
	\includegraphics[width=\linewidth]{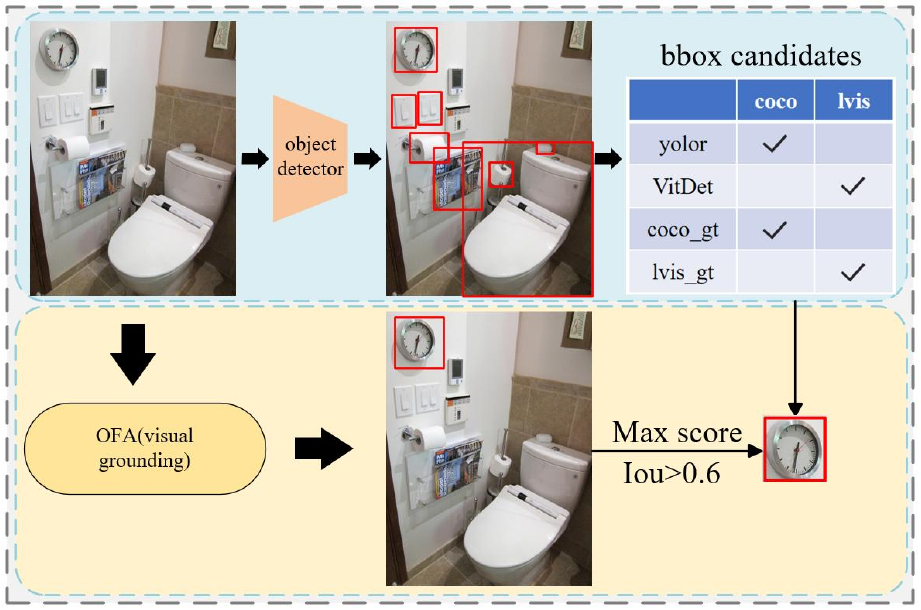}
	\caption{Postprocessing Stage.}
\label{fig: similarity}
\end{figure}

\subsection{Postprocessing Stage}
The third part is the post-processing stage, the bounding box position predicted by OFA is roughly correct, but the accuracy of coordinates is not higher than object detector. Therefore, by calculating the iou, we replace the predicted bounding box with the object detector's bounding box. For each image, all bounding boxes in the image are detected by object detectors yolor and vitDet, and sorted according to the confidence, which are called candidate Bounding boxes. Then, for the predicted bounding box, the first candidate bounding box with iou higher than zero point six is selected to replace the predicted bounding box. Finally, the model ensemble includes three ways, the first is five folds with standard Corase tuning, the second is ten folds with standard Corase tuning, and the third is five folds with back translation Corase tuning. The back translation was applied to the coarse tuning stage, which expanded the diversity of question data.
\section{Experiments}

\subsection{Dataset.}
The Toloka dataset comprises images paired with textual questions, where each entry includes a question-image pair annotated with ground truth bounding box coordinates pinpointing the visual answer. In total, the dataset consists of 45,199 instances distributed across three subsets: 38,990 instances in the training set, 1,705 instances in the public test set, and 4,504 instances in the private test set.

The dataset is structured with several key columns: "image" contains URLs linking to images hosted on a public content delivery network; "question" provides English-language queries associated with each image. Additional metadata includes "width" and "height" integers representing the dimensions of each image. For bounding box annotation, the dataset includes "left," "top," "right," and "bottom" integers detailing the coordinates that define the spatial extent of the object or region in the image that corresponds to the answer to the question.

\subsection{Leadboards.}

\begin{table}[!htbp]
\centering
\renewcommand{\arraystretch}{1.3} % Adjusts the row height
\setlength{\tabcolsep}{10pt} % Adjusts the column width
\begin{tabular}{c|c}
\toprule[0.8pt]
{ Method} & { Score} \\ \hline
Baseline & 71.0 \\
Pseudo Answer & 73.5 \\
Template & 74.2 \\
Coarse Tuning & 75.1 \\
Postprocessing & 75.8 \\
Test Public & 76.5 \\
Test Private & 76.342 \\ 
\bottomrule[0.8pt]
\end{tabular}
\caption{Results of each component.}
\label{tab: result}
\end{table}

\noindent Table~\ref{tab: result} shows the improvement in model performance by each of our components. The baseline is defined as the OFA model which directly inputs the competition dataset and reasoning. The performance of coarse tuning and pseudo answer increased the most because the pseudo answer showed the category of objects corresponding to the bounding box. On the test public set, our method obtained a seventy-six point five score, and on the test private set, our method obtained a seventy-six point three score, which shows that our method has strong generalization.

{
    \small
    \bibliographystyle{ieeenat_fullname}
    \bibliography{main}
}


\end{document}